\documentclass{article}
\usepackage{spconf,amsmath,graphicx,hyperref}
\usepackage{amssymb}
\usepackage{subcaption}
\usepackage{booktabs}
\usepackage{graphicx} 

\title{Refine-Control: A Semi-Supervised Distillation Method for Conditional Image Generation}

\name{Yicheng Jiang$^{1}$, Jin Yuan$^{2}$, Hua Yuan$^{1}$, Yao Zhang$^{2}$, Yong Rui$^{1,\dagger}$}
\address{ $^{1}$ School of Computer Science and Engineering, Southeast University, Nanjing, China \\
      $^{2}$AI Lab, Lenovo Research, Beijing, China \\
      $\{$ycjiang, yuanhua$\}$@seu.edu.cn, $\{$yuanjin1, zhangyao22$\}$@lenovo.com, yongrui@outlook.com}
\begin{document}
%
\maketitle
\begin{abstract}



Conditional image generation models have achieved remarkable results by 
leveraging text-based control to generate customized images.
However, the high resource demands of these models and the scarcity of well-annotated data have hindered their deployment on edge devices,
leading to enormous costs and privacy concerns, especially when user data is sent
to a third party.
To overcome these challenges, we propose Refine-Control, a semi-supervised distillation framework.
Specifically, we improve the performance of the student model by introducing a tri-level knowledge fusion loss
to transfer different levels of knowledge.
To enhance generalization and alleviate dataset scarcity, we introduce a semi-supervised 
distillation method utilizing both labeled and unlabeled data.
Our experiments reveal that Refine-Control achieves significant reductions in computational cost and
latency, while maintaining high-fidelity generation capabilities and controllability, 
as quantified by comparative metrics.

\end{abstract}
\begin{keywords}
Model Distillation, Diffusion Models, Conditional Control Networks, Image Generation
\end{keywords}
\section{Introduction}
\label{sec:intro}

Recently, researchers have applied controllable modules to guide text-to-image diffusion models, 
such as ControlNet \cite{zhang2023adding}
to achieve fine-grained control over image generation.
As a controllable add-on module to the foundational
model, it encodes extra spatial conditions into fine-grained guidance signals. 
This allows for pixel-level control without compromising the powerful generative capabilities of 
the pre-trained model, leading to its widespread adoption in various applications \cite{zhong2025sketch2anim,wang2025lavie,xu2025ootdiffusion}, 
including image editing \cite{feng2025dit4edit,huang2025pfb}.

While this ``foundation + module'' paradigm has significantly enhanced controllability, 
its success has also introduced severe challenges:
1) High Computational and Memory Costs. The architecture of add-on modules like ControlNet is intricately integrated 
with the underlying diffusion model. Typically, the size of these modules exceeds half of the model's overall structure \cite{SD3-controlnet-inpainting,flux2024}.
This substantial resource requirement severely hampers the 
deployment and application of advanced controllable generative models on edge devices.
2) Reliance on Large-Scale Finely Annotated Data. Training a high-performance controllable module 
often requires massive, high-quality datasets. Acquiring such datasets is expensive and time-consuming,
especially in tasks such as image inpainting.
3) Structural Hurdles in Knowledge Distillation. While model distillation is an effective
approach for model compression \cite{chadebec2025flash,hu2024snapgen,li2023snapfusion}, 
the significant architectural mismatches between teacher and student
ControlNet models pose unique challenges for its direct application.

To address these challenges, we propose Refine-Control, a novel two-stage semi-supervised distillation framework. 
The motivation behind our two-stage approach is to decouple the distillation process into 
foundational mapping and advanced refinement.
In the first stage,
we leverage a fully annotated dataset for supervised learning, enabling the student model 
to quickly learn the initial mapping from conditions to outputs. 
In the second stage, we use an easily accessible dataset with unlabeled data for self-supervised fine-tuning. 
This improves the generalization and generation quality of the student model.
Considering mainstream loss functions are often insufficient to deal with the structural mismatches of ControlNets, 
we design a specialized tri-level knowledge fusion loss 
to transfer hierarchical knowledge of the teacher.
We also utilize local prompts to alleviate the misinterpretation problems caused by global prompts, 
when facing scenarios with multiple objects similar in semantics, thereby enhancing controllability in complex scenes.

In summary, our contributions are as follows:
\begin{itemize}
  \item  We propose a novel \textbf{two-stage distillation framework} enabling efficient knowledge transfer
  even with ``data challenges".
   
  \item We introduce a specialized \textbf{tri-level knowledge fusion loss} to transfer hierarchical knowledge
  during distillation.
  
  \item A novel data pipeline with local prompts is introduced to enhance controllability.

\end{itemize}

\section{The Refine-Control Framework} 
\label{sec:method}

We frame the image-inpainting distillation as a two-stage process:
foundational learning and advanced refinement.
Our approach is designed to systematically build the student model's capabilities by decoupling 
the training process into these two distinct and complementary phases.
The primary goal of the first stage is to establish a robust baseline understanding of the task, and to
ensure the student can produce structurally coherent results.
Subsequently we apply self-supervised fine-tuning
in the second stage. This enables the student to further learn
from the teacher's generative logic, enhancing its generalization and generation quality.

\subsection{The Supervised Distillation Stage}
To enable the student model to quickly learn the initial mapping from conditions to outputs in this stage, 
we design a tri-level knowledge fusion loss that effectively transfers hierarchical knowledge 
from the teacher to the student.
It contains three different components: a goal-oriented task loss ($L_{task}$), 
an output-oriented distillation loss ($L_{distill}$), and a process-oriented asymmetric feature loss ($L_{af}$).
This work primarily focuses on image inpainting tasks, and the total loss for this stage is:

\begin{equation}\label{eq:loss in 1st stage} 
  L_{stage1} = \lambda_{task} \cdot L_{task} +\lambda_{distill} \cdot L_{distill} + \lambda_{af} \cdot L_{af},
\end{equation}
where $\lambda_{task}$, $\lambda_{distill}$, and $\lambda_{af}$ are weighting hyperparameters.

\begin{figure}[ht]
    \centering
    \includegraphics[width=0.9\linewidth]{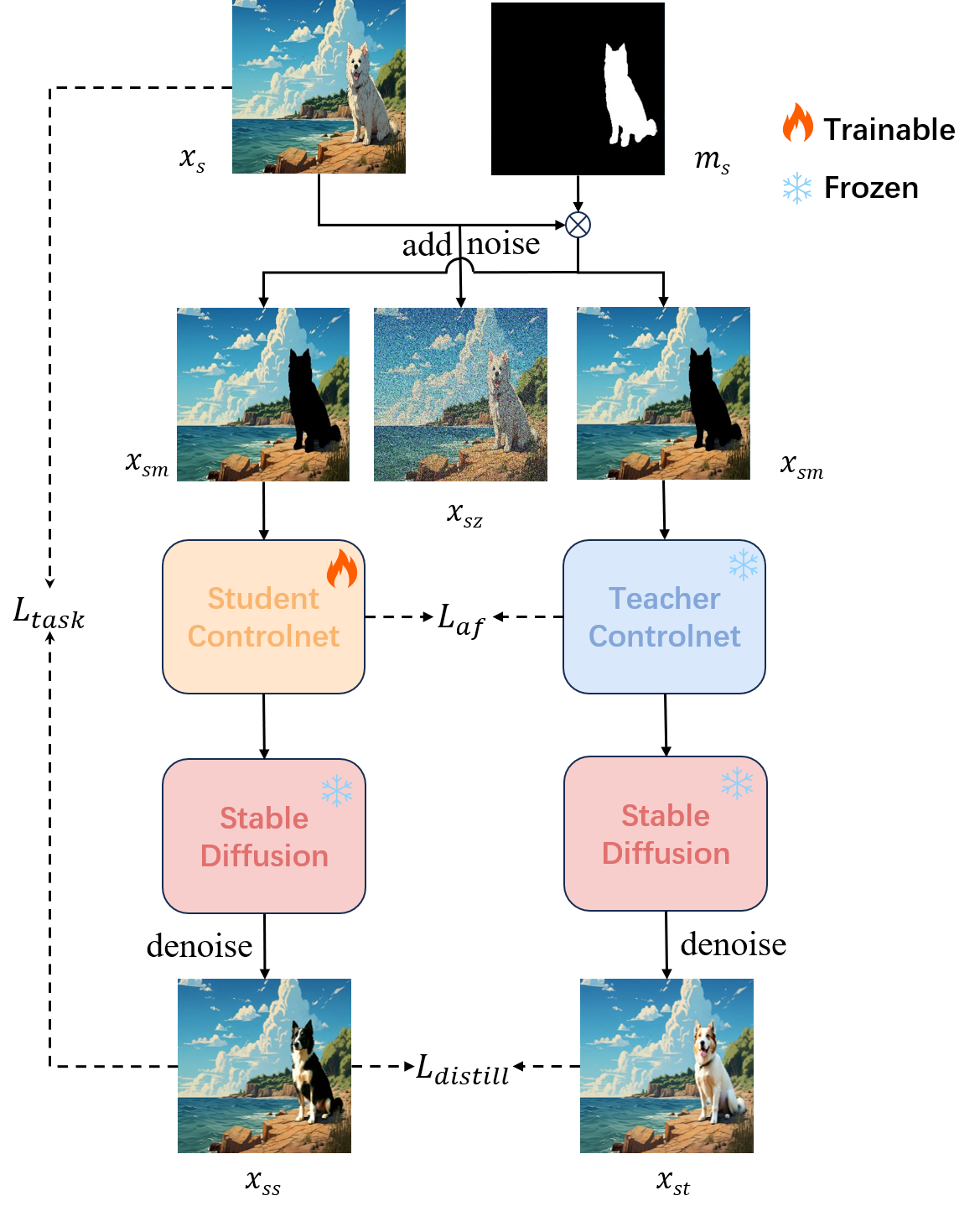}
    \captionof{figure}
      {
        Illustration of tri-level knowledge fusion loss in the supervised distillation stage.
        The student is optimized via three complementary losses: task-specific performance, 
        final output mimicry, and intermediate feature representation. 
        The Stable Diffusion backbone remains frozen.
      }
    \label{fig:first loss}
    
\end{figure}

\subsubsection{Mask-Weighted Task Loss}

Instead of computing the loss over the entire image, we guide the model to focus on the 
user-specified inpainting region, which represents the key area of interest. To achieve this, 
we introduce a mask-weighted task loss. Given the binary mask $M$ (where inpainting regions are 1), 
we define a weight matrix $\omega_{mask}$:

\begin{equation}\label{eq:mask_loss}
  \omega_{mask} = \alpha \cdot M + (1-\alpha) \cdot (J-M),
\end{equation}
where $J$ is a matrix of ones with the same shape as $M$, and $\alpha > 0.5$ is a hyperparameter that 
assigns higher importance to the masked areas.
The task loss is then defined based on the Flow Matching objective \cite{lipman2022flow}:

\begin{equation}\label{eq:task_loss}
  L_{task} = \mathbb{E}_{x_0, \epsilon, t} \left[ ||\omega_{mask} \odot (\epsilon - \epsilon_\theta(x_t, t, c))||^2 \right],
\end{equation}
where $c$ denotes the generation constraints, including text prompts $c_{t}$ and ControlNet constraints 
$c_{control}$, $x_0$ represents the original image, $\epsilon$ denotes the noise, and $t$ signifies the diffusion steps. $x_t$ is the noisy sample obtained by adding noise $\epsilon$ to $x_0$ at timestep $t$, and $\epsilon_\theta$ is the final output of the model.
This formulation ensures that errors within the inpainting region are penalized more heavily, 
guiding the student model to prioritize generation quality where it matters most.

\subsubsection{Distillation Loss}

Following the success of knowledge distillation in diffusion models \cite{chadebec2025flash,hu2024snapgen,li2023snapfusion}, we employ an output-level distillation loss.
This loss encourages the student model $\epsilon_\theta$ to replicate the final denoising output of the 
teacher model $\epsilon_{T}$, effectively transferring its overall generative capability. 
The loss is defined as:

\begin{equation}\label{l_noise}
 L_{distill} = \mathbb{E}_{x_0, \epsilon, t} \left[ ||\epsilon_{T}(x_t, t, c) - \epsilon_\theta(x_t, t, c)||^2_2 \right],
\end{equation}

\subsubsection{Asymmetric Feature Loss}

A key challenge in distilling ControlNet is the architectural mismatch, as our student model has 
fewer layers ($S=12$) than the teacher model ($T=23$). To bridge this structural gap, we propose a 
\textbf{many-to-one feature alignment} and a \textbf{one-to-many injection} strategy.
Instead of a simple one-to-one mapping, each layer of the student network is designed to learn the 
aggregated features from two corresponding layers of the teacher network, 
and to connect to the fundamental model layers to which its corresponding group of teacher layers would be injected.

Specifically, let $f_S^j$ be the output feature map of the j-th layer of the student ControlNet, 
and $f_T^i$ be that of the i-th layer of the teacher. 
Our alignment strategy ensures that the j-th student layer approximates the features of the 
(2j-1)-th and (2j)-th teacher layers, 
where $f_S^j$ would be injected into the fundamental model layer to which the teacher layers (2j-1)-th and (2j)-th are connected.
This ensures a more comprehensive feature transfer across different semantic levels. 
The asymmetric feature loss is formulated as:

\begin{equation}
  \label{eq:cf_loss}
  L_{af} = E_{x,c_t,m} [ \sum_{j=1}^{S} || 2 \cdot f_S^j(\cdot) - (f_T^{2j-1}(\cdot) + f_T^{2j}(\cdot)) ||^2 ],
\end{equation}
where the parameters of the teacher model are frozen.
This loss directly optimizes the student's intermediate representations, 
which is crucial for preserving the fine-grained control capabilities of the teacher.

\subsection{Self-Supervised Fine-tuning Stage}
Considering the scarcity of finely labeled data, we introduce a self-supervised learning stage 
in the second phase.
The goal of this phase is to refine the model's internal generative logic, prompting it to rely solely 
on the teacher's guidance rather than ground-truth references. 
This mitigates potential overfitting to the training data and improves the coherence of generated content.
Meanwhile, this stage also aims to increase the practicality of the distillation scheme, 
demonstrating that the student model can be trained on partial data, 
which is more accessible in real-world scenarios.
In this stage, we apply a bi-level knowledge fusion loss  combining the final output mimicry loss 
$L_{distill}$ and intermediate feature representation loss $L_{af}$ shown in Fig.\ref{fig:first loss}.
The optimization objective is defined as

\begin{equation}
  L_{stage2} = \lambda_{distill} \cdot L_{distill} + \lambda_{af} \cdot L_{af}.
\end{equation}


\section{Experiments}
\label{sec:exp}

\subsection{Experimental Setup}

\subsubsection{Datasets}
We construct our datasets from  LAION-2B \cite{schuhmann2022laion},
using the Segment Anything Model (SAM) \cite{kirillov2023segany} for semantic segmentation.
The supervised dataset, $D_{sl}$, contains 130k triplets of (image, mask, local prompt). 
The self-supervised dataset, $D_{ssl}$, contains another 130k pairs of (masked image, local prompt).
To improve generation precision, we utilize Qwen2.5-VL \cite{Qwen2.5-VL}, supplemented with human annotation,
to generate local prompts only describing the masked regions, aiming to mitigate potential ambiguities
caused by global prompts, as shown in Fig.\ref{fig:local prompt}. 
For evaluation, we create a testing dataset $D_{test}$ from 
Unsplash \cite{kulshreshtha2022feature} in a similar way.

\subsubsection{Implementation Details}
We adopt SD3-Inpainting-ControlNet \cite{SD3-controlnet-inpainting} as our teacher model, 
which is a 23-layer ControlNet model.
The student model is designed to have 12 layers, initialized with the first 12 layers of 
SD3-medium \cite{esser2024scaling}, which is the fundamental model.

\subsection{Quantitative Results}

We compare our proposed distillation method against two critical baselines: 
(1) the original teacher model, SD3-Inpainting-ControlNet, 
(2) a student model with the same architecture trained from scratch ($Student_{fc}$).
In $Student_{ours}$, we set the hyperparameters
$\lambda_{task}=4$, $\lambda_{distill}=\lambda_{af}=1$ in Eq.\ref{eq:loss in 1st stage} and $\alpha=0.975$ in Eq.\ref{eq:mask_loss}.

\begin{table}[h]
  \centering
  \caption{Distillation evaluation.}
  \label{tab:ex-eval}
  \small
  \resizebox{\linewidth}{!}{
  \begin{tabular}{cccccc}
    \toprule
    Model & PSNR$\uparrow$ & SSIM$\uparrow$ & FID$\downarrow$ & LPIPS$\downarrow$ & CMMD$\downarrow$ \\
    \midrule

    SD3-Inpainting & $\boldsymbol{19.7083}$ & $\boldsymbol{0.7535}$ & $\boldsymbol{40.80}$ & $\boldsymbol{0.1344}$ &  $\boldsymbol{0.0321}$ \\
    $Student_{fc}$ & 17.7784 & 0.7373 & 48.15 & 0.1593 & 0.0672 \\ 
    $Student_{ours}$ & $\underline{19.2546}$ & $\underline{0.7505}$ & $\underline{43.32}$ & $\underline{0.1379}$ & $\underline{0.0342}$  \\

    \bottomrule
  \end{tabular}
  }
\end{table}

In Table \ref{tab:ex-eval}, we present the results in terms of PSNR \cite{wang2004image},
SSIM \cite{wang2004image}, FID \cite{heusel2017gans}, LPIPS \cite{zhang2018unreasonable},
and CMMD \cite{jayasumana2024rethinking}.
Here, $\uparrow$ indicates higher is better, and 
$\downarrow$ indicates lower is better. 
The best results for each metric are highlighted in bold, and the second-best results are underlined.

It can be seen that $Student_{ours}$ achieves comparable results to the teacher model,
which is twice its size,
and outperforms $Student_{fc}$ significantly in all metrics, demonstrating that
distillation is a far more effective approach than simple training for 
creating high-quality, compact models, thereby validating the effectiveness of our knowledge transfer framework.

We also compare $Student_{ours}$ with other distillation methods and inpainting models.
The compared models include Dreamshaper-v8 \cite{dreamshaper}, SD2-Inpainting \cite{rombach2022high}, SD3-Inpainting-ControlNet,
FLUX-Inpainting-ControlNet \cite{flux2024}, and $Student_{ssd}$, a model using the distillation method proposed 
in \cite{gupta2024progressive}. All baselines except $Student_{ssd}$ use official implementations.

\begin{table}[h]
  \centering
  \caption{Quantitative results of previous distillation methods and generative models in image inpainting tasks.}
  \label{tab:baseline}
  \small
  \resizebox{\linewidth}{!}{
  \begin{tabular}{cccccc}
    \toprule
    Model & PSNR$\uparrow$ & SSIM$\uparrow$ & FID$\downarrow$ & LPIPS$\downarrow$ & CMMD$\downarrow$\\
    \midrule

    Dreamshaper-v8 \cite{dreamshaper} & 18.3507 & 0.682 & 55.10 & 0.1764 & 0.0898  \\ 
    SD2-Inpainting \cite{rombach2022high} & 19.0982 & 0.6889 & 44.63 & 0.1514 & 0.0442\\
    SD3-Inpainting \cite{SD3-controlnet-inpainting} & $\underline{19.7083}$ & $\underline{0.7535}$ & $\underline{40.80}$ & $\underline{0.1344}$ &  $\boldsymbol{0.0321}$ \\
    FLUX-Inpainting \cite{flux2024} & $\boldsymbol{20.5}$  &	$\boldsymbol{0.7734}$	& $\boldsymbol{39.62}$ & $\boldsymbol{0.1272}$ & 0.0604 \\
    $Student_{ssd}$ & 18.6984 & 0.7478 & 44.52 & 0.1453 & 0.0484 \\
    $Student_{ours}$ & 19.2546 & 0.7505 & 43.32 & 0.1379 & $\underline{0.0342}$  \\

    \bottomrule
  \end{tabular}
  }
\end{table}

As shown in Table \ref{tab:baseline}, our 12-layer $Student_{ours}$ achieves comparable results to these baselines.
FLUX-Inpainting-ControlNet and 23-layer SD3-Inpainting-ControlNet outperform other models, 
but they require significantly larger computational costs and latency than $Student_{ours}$.
    
    

\subsection{Ablation Study}
To validate the effectiveness of different modules in our proposed framework, we perform ablation
studies. Specifically, we designed several variant models, 
each omitting a different component to evaluate its specific contribution to
the overall performance.

\begin{itemize}
  \item $Student_{wo\_ mask}$: The variant removes the mask region weighting in the task loss $L_{task}$.
  \item $Student_{one\_stage}$: The variant removes the two-stage distillation, combining the training 
  datasets $D_{sl}$ and $D_{ssl}$ randomly in a single training stage.
  \item $Student_{sl}$: The variant is trained solely with supervised learning on the $D_{sl}$ dataset, 
  while maintaining the same total training steps as the two-stage approach.
  \item $Student_{wo\_cf}$: The model removes the asymmetric feature loss $L_{af}$.
\end{itemize}

\begin{table}[h]
  \centering
  \caption{Ablation studies.}
  \label{tab:消融实验}
  \small
  \resizebox{\linewidth}{!}{
  \begin{tabular}{cccccc}
    \toprule
    Model & PSNR$\uparrow$ & SSIM$\uparrow$ & FID$\downarrow$ & LPIPS$\downarrow$ & CMMD$\downarrow$\\
    \midrule

    $Student_{wo\_mask}$ & 18.9841 & 0.7484 & 43.98  & 0.1432 & 0.0360 \\
    $Student_{one\_stage}$ & 18.8139 & 0.7471 & 45.53 & 0.1448 & 0.0470\\
    $Student_{sl}$ & 18.7546 & 0.7468 & 44.26 & 0.1438 & 0.0446\\
    $Student_{wo\_cf}$ & 19.1739 & 0.7500 & 43.64 & 0.1399 & 0.0368\\
    $Student_{ours}$ & $\boldsymbol{19.2546}$ & $\boldsymbol{0.7505}$ & $\boldsymbol{43.32}$ & $\boldsymbol{0.1379}$ & $\boldsymbol{0.0342}$ \\

    \bottomrule
  \end{tabular}
  }
\end{table}

As shown in Table \ref{tab:消融实验}, the removal of the mask region weighting, the two-stage distillation pipeline,
the second stage self-supervised learning, or the asymmetric feature loss
all consistently decrease in the performance of the student model.
This demonstrates the effectiveness of each component in our proposed distillation framework.

\subsection{Qualitative Analysis }

\begin{figure}[!htbp]
  \centering
  \begin{minipage}{0.05\linewidth} 
    
    \centerline{\rotatebox{90}{Global}}
    \vspace{30pt}
    \centerline{\rotatebox{90}{Local}}
    \vspace{10pt}
  \end{minipage}
  \begin{minipage}{0.25\linewidth}
    \vspace{3pt}
    \centerline{\includegraphics[width=\textwidth]{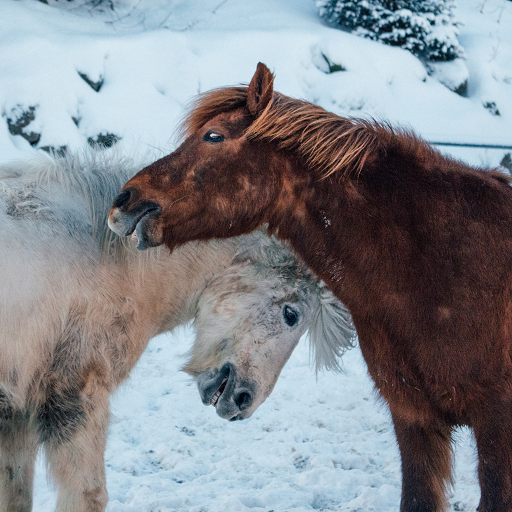}}
    \vspace{3pt}
    \centerline{\includegraphics[width=\textwidth]{horse_image_low.png}}
    \centerline{\small Original image}
  \end{minipage}
  \begin{minipage}{0.25\linewidth}
   \vspace{3pt}
   \centerline{\includegraphics[width=\textwidth]{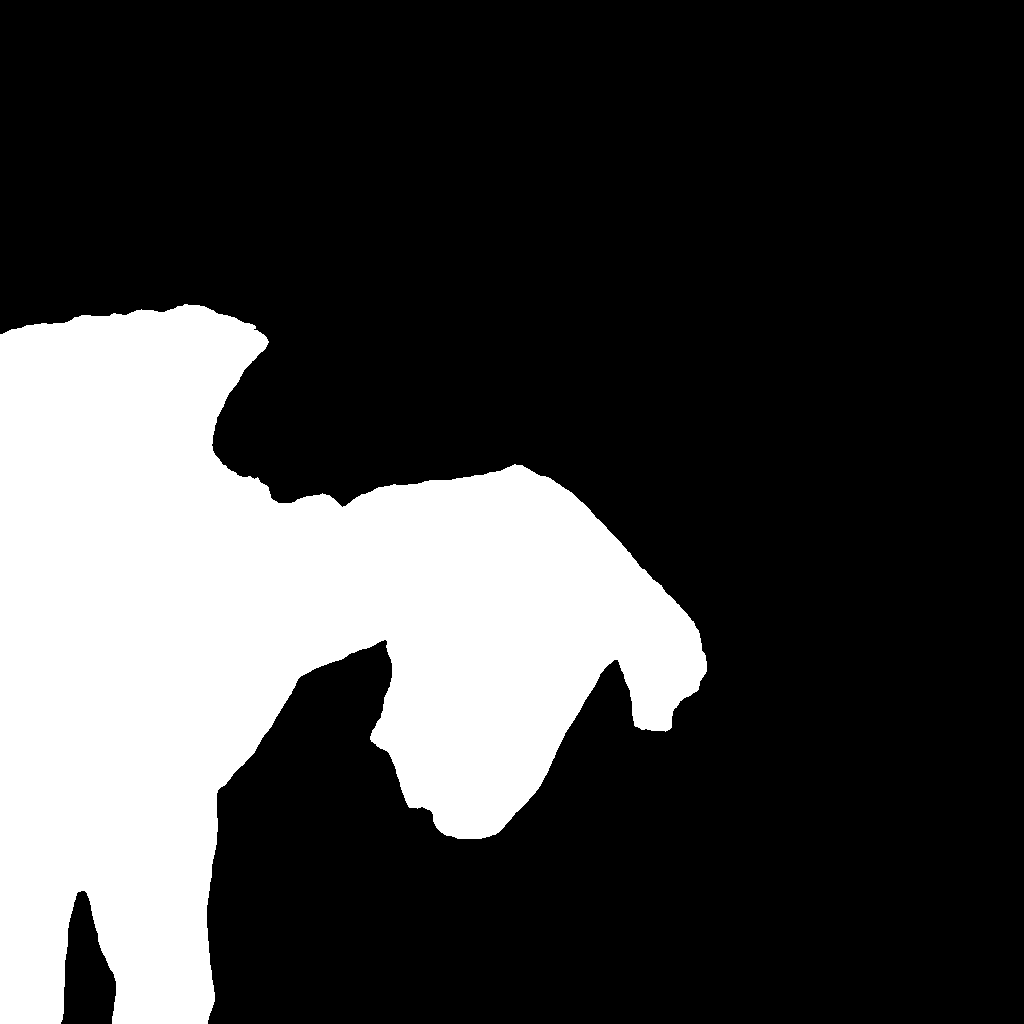}}
   \vspace{3pt}
   \centerline{\includegraphics[width=\textwidth]{horse_mask.png}} 
   \centerline{\small Mask}
 \end{minipage}
 \begin{minipage}{0.25\linewidth}
   \vspace{3pt}
   \centerline{\includegraphics[width=\textwidth]{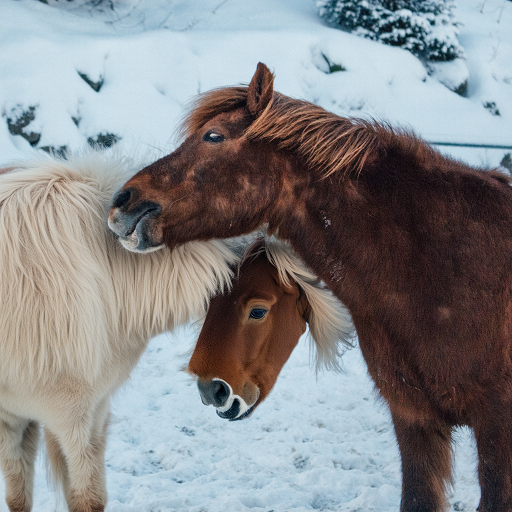}}
   \vspace{3pt}
   \centerline{\includegraphics[width=\textwidth]{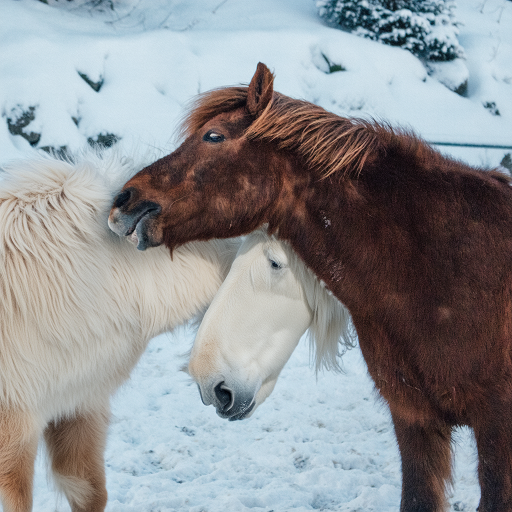}}
   \centerline{\small Generation}
 \end{minipage}
 \caption{Comparison of generation results with global and local prompts. The introduction of local prompts alleviates the misunderstanding of the generation task.}
 \label{fig:local prompt}
\end{figure}

We demonstrate the robustness of our distilled model in alleviating misinterpretations
when handling multiple semantically similar target objects in image inpainting tasks.
As shown in Fig.\ref{fig:local prompt}, when users only want to inpaint a specific horse,
by using a local prompt, ``one white fluffy pony", our model
can accurately generate the content in the specified region, outperforming SD3-Inpainting-ControlNet
using global prompt, ``Two fluffy ponies, one brown and one white, nuzzling each other in the snow",
which leads to the misinterpretation of the generation task.

\begin{figure}[!htbp]
  \centering
  \begin{minipage}{0.22\linewidth}
    \vspace{3pt}
    \centerline{\includegraphics[width=\textwidth]{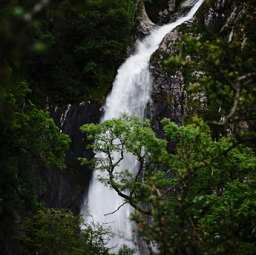}}
    \vspace{3pt}
    \centerline{\includegraphics[width=\textwidth]{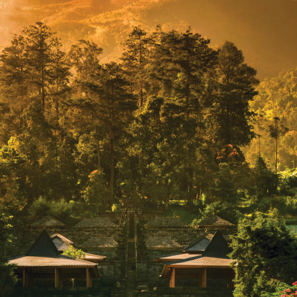}}
    \vspace{3pt}
    \centerline{\includegraphics[width=\textwidth]{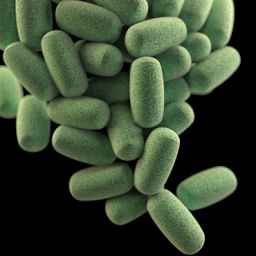}}
    \vspace{3pt}
    \centerline{\small Original image}
  \end{minipage}
  \begin{minipage}{0.22\linewidth}
    \vspace{3pt}
    \centerline{\includegraphics[width=\textwidth]{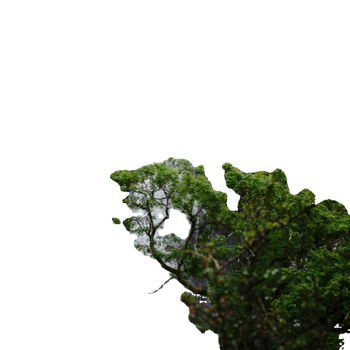}}
    \vspace{3pt}
    \centerline{\includegraphics[width=\textwidth]{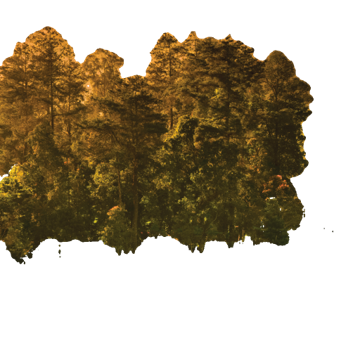}}
    \vspace{3pt}
    \centerline{\includegraphics[width=\textwidth]{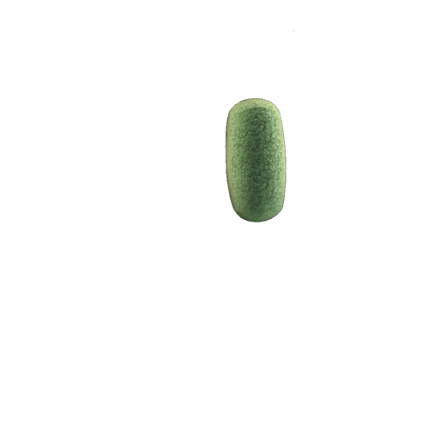}}
    \vspace{3pt}
    \centerline{\small Masked image}
  \end{minipage}
  \begin{minipage}{0.22\linewidth}
   \vspace{3pt}
   \centerline{\includegraphics[width=\textwidth]{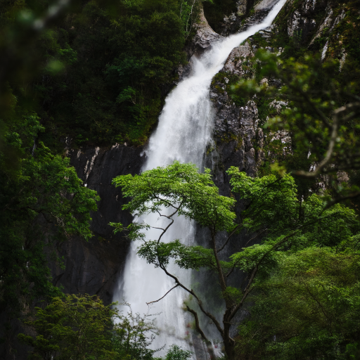}}
   \vspace{3pt}
   \centerline{\includegraphics[width=\textwidth]{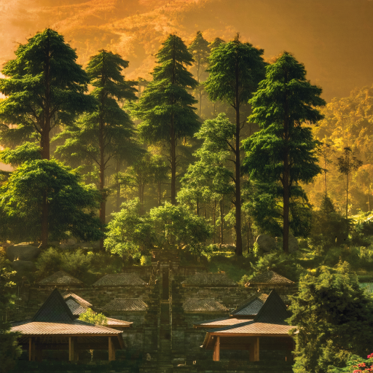}}
   \vspace{3pt}
   \centerline{\includegraphics[width=\textwidth]{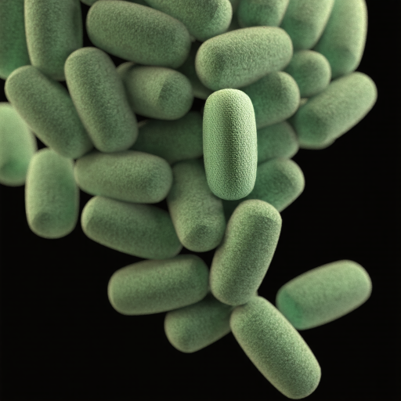}}
   \vspace{3pt}
   \centerline{\small $Student_{ours}$}
 \end{minipage}
 \begin{minipage}{0.22\linewidth}
   \vspace{3pt}
   \centerline{\includegraphics[width=\textwidth]{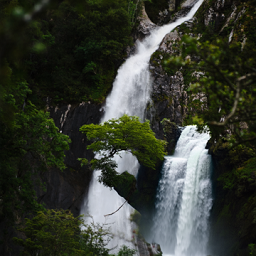}}
   \vspace{3pt}
   \centerline{\includegraphics[width=\textwidth]{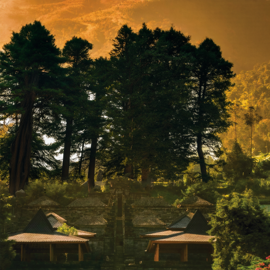}}
   \vspace{3pt}
   \centerline{\includegraphics[width=\textwidth]{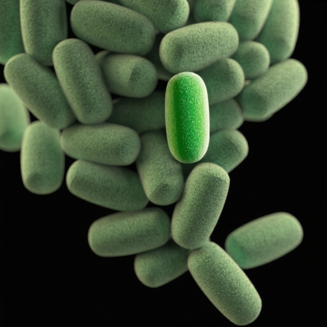}}
   \vspace{3pt}
   \centerline{\small $Student_{fc}$}
 \end{minipage}
 \caption{Qualitative Analysis of $Student_{ours}$ and $Student_{fc}$. $Student_{ours}$ outperforms $Student_{fc}$ in both quality and coherence.}
 \label{fig:生成结果的视觉比较}
\end{figure}

Furthermore, we provide a visual comparison of different models.
As shown in Fig.\ref{fig:生成结果的视觉比较}, $Student_{ours}$ effectively captures the information in unmasked areas without being misled by it.
It is evident that $Student_{ours}$ produces results that are far superior in quality and coherence 
to the model trained from scratch, $Student_{fc}$.

\section{Conclusion}
\label{sec:Conclusion}

This paper proposed a novel and effective distillation method  
for conditional diffusion models, specifically tailored for inpainting ControlNet,
using a two-stage distillation strategy.
Through the use of a tri-level knowledge fusion loss that combines distinct levels of knowledge transfer,  
and a second self-supervised fine-tuning stage, 
this method effectively reduces the computational requirements for ControlNet while 
preserving its high-fidelity generative and control capabilities.

\newpage
\small
\bibliographystyle{IEEEbib}
\bibliography{strings,refs}

\end{document}